\pdfoutput=1

\documentclass[11pt]{article}

\usepackage{ACL2023}

\usepackage{times}
\usepackage{latexsym}

\usepackage[T1]{fontenc}

\usepackage[utf8]{inputenc}

\usepackage{microtype}

\usepackage{inconsolata}

\usepackage{booktabs} 

\usepackage{float}

\usepackage{graphicx}

\usepackage{tabularx}

\newcommand{\smallfootnote}[1]{\footnote{\tiny #1}}

\title{LLMs as mirrors of societal moral standards: reflection of cultural divergence and agreement across ethical topics}

\author{Mijntje Meijer, Hadi Mohammadi \and Ayoub Bagheri \\
        Department of Methodology and Statistics, Utrecht University, Padualaan 14, Utrecht, The Netherlands \\
        \texttt{y.f.s.s.meijer@students.uu.nl, h.mohammadi@uu.nl, a.bagheri@uu.nl}}

\begin{document}
\maketitle

\begin{abstract}
Large language models (LLMs) have become increasingly pivotal in various domains due the recent advancements in their performance capabilities. However, concerns persist regarding biases in LLMs, including gender, racial, and cultural biases derived from their training data. These biases raise critical questions about the ethical deployment and societal impact of LLMs. Acknowledging these concerns, this study investigates whether LLMs accurately reflect cross-cultural variations and similarities in moral perspectives. In assessing whether the chosen LLMs capture patterns of divergence and agreement on moral topics across cultures, three main methods are employed: (1) comparison of model-generated and survey-based moral score variances, (2) cluster alignment analysis to evaluate the correspondence between country clusters derived from model-generated moral scores and those derived from survey data, and (3) probing LLMs with direct comparative prompts. All three methods involve the use of systematic prompts and token pairs designed to assess how well LLMs understand and reflect cultural variations in moral attitudes. The findings of this study indicate overall variable and low performance in reflecting cross-cultural differences and similarities in moral values across the models tested, highlighting the necessity for improving models' accuracy in capturing these nuances effectively. The insights gained from this study aim to inform discussions on the ethical development and deployment of LLMs in global contexts, emphasizing the importance of mitigating biases and promoting fair representation across diverse cultural perspectives. \end{abstract}

\section{Introduction}

Over the past few years, large language models (LLMs) have become increasingly prominent in current discussions, both in the scientific and public realm \citep{Bender:2021}. Due to significant advances in model performance, LLMs now offer promising avenues for applications across a wide range of fields. For instance, large language models are increasingly being used in various applications that impact people's daily lives profoundly, such as search engines, recommendation systems, and automated decision-making systems. However, while the recent performance of LLMs, such as OpenAI's newly released GPT-4, is impressive, there are also areas of concern. An important concern regarding LLMs is whether and in which areas these models may exhibit bias, such as gender, racial and cultural bias. \\
\indent Large language models are sensitive to embedded bias due to the way they are trained, as they pick up the societal and cultural biases present in the training data. Due to the fact that LLMs are trained on large amounts of data, the models may be (partially) trained on data that reflects embedded societal and cultural prejudices, such as news articles or social media posts \citep{karpouzis2024, mishra2024}. Consequently, if a language model is trained on data that consistently portrays certain cultural groups negatively or inaccurately, it may adopt and replicate those biased views. Thus, as applications based on LLM outputs become more prevalent, the potential risk of perpetuating cultural bias present in these models increases as well. Therefore, it is important to assess whether LLMs accurately reflect the empirically observed moral judgments present in different cultures. Despite its importance, this issue remains understudied in the literature \citep{Arora:2022, liu2023}. In evaluating whether LLM’s faithfully capture an understanding of the moral judgments across different cultures in a broad sense, it is crucial to assess how well these models reflect inter-cultural differences and similarities in moral judgments. This can be captured in the following research question: "To what extent do language models capture cultural diversity and common tendencies regarding topics on which people around the world tend to diverge or agree in their moral judgments?". \\
\indent Assessing how well LLMs reflect the differences and similarities across cultures on various moral topics will contribute to developing a wider understanding of how accurately these models grasp empirically observed cultural moral values, such as those recorded in surveys or historical data. As such, this research adds to the scientific debate regarding LLMs’ understanding of cross-cultural moral values, which, as mentioned above, has been underexposed. \citet{Benkler:2023} emphasize the value of testing and comparing a variety of LLMs, since previous works have shown subtle differences in the ways different LLMs generate output and may embed bias. Moreover, the research question holds significant societal relevance: as the use of LLMs becomes more prevalent in all spheres of life, it is important to assess whether these models accurately represent the diversity of cultural perspectives and moral judgments across the globe \citep{liu2023}. If this is not the case, model output may perpetuate bias, prejudice and unfairness by representing cultural differences and similarities on moral topics among groups inaccurately. LLMs that do accurately capture the differences and similarities regarding moral judgment across cultural groups, on the other hand, can help identify shared values and ethical principles across diverse communities, thereby aiding in fostering cross-cultural understanding and collaboration. \\
\indent In short, assessing whether LLMs reflect cultural diversity and  common tendencies regarding moral topics adds to the scientific debate, and holds significant societal value due to its implications for minimizing bias and prejudice, promoting accurate depictions of cultural groups and cultivating cross-cultural understanding. The aim of this study is to provide insights into the capabilities and societal implications of large language models, specifically related to their understanding of cross-cultural moral values. The study employs three main methods in evaluating models' abilities in capturing and replicating cross-cultural variations and similarities in moral perspectives: (1) comparing variances of model-generated and survey-based moral scores across countries, (2) assessing alignment between country clusters derived from model-generated moral scores and those derived from survey-based moral scores, and (3) probing LLMs with direct comparative prompts to evaluate their recognition of cultural differences and similarities in moral judgments. By assessing LLMs' capabilities in understanding and reflecting cross-cultural differences and similarities in moral perspectives through these methods,  
the insights gained from this study significantly contribute to informed discussions on the ethical deployment of LLMs.

This paper is structured as follows: first, an overview of related literature will be provided. Next, the data and methodology used in this study will be described. Then, the results are presented, followed by the discussion and conclusion.

\section{Literature review}

\subsection{Cross-cultural understanding of moral judgments in LLMs}

Moral judgments refer to evaluations of certain actions, intentions, and individuals somewhere along a spectrum of ‘good’ or ‘bad’. These judgments can significantly vary across different cultures, influenced by factors such as religion, societal norms, and historical contexts (\citet{Haidt:2001, Shweder:1997}). As highlighted by \citet{graham2016}, Western, Educated, Industrialized, Rich and Democratic (W.E.I.R.D.) cultures are generally more inclined to endorse moral codes that emphasize individual rights and independence, while non-W.E.I.R.D. cultures tend to more strongly emphasize duty-based communal obligations and spiritual purity. This leads people in W.E.I.R.D. (autonomy-endorsing) cultures to view personal actions such as sexual behaviors as a matter of individual rights, while those in non-W.E.I.R.D. (community-endorsing) cultures are inclined to perceive them as a collective moral concern. \citet{Johnson:22} point out that, while there are many resonant and overlapping values amongst the world's cultures, there are also many conflicting yet equally valid values. \citet{Johnson:22} and 
 \citet{Benkler:2023} refer to this notion as 'moral value pluralism' and underscore its importance. \citet{kharchenko2024} emphasize LLMs' limitations in accurately representing moral value pluralism, stating that general values become improperly embedded in transformer driven models due to the lack of diversity in training data. \citet{du-2024} also state that the emphasis of English in LLMs' training data overshadows the linguistic diversity inherent to human languages and limits the scope of model applicability and innovation. The authors highlight that it is therefore important for LLMs to be trained on multilingual data, and mention that larger data volumes and bigger model sizes enhance performance as well. \citet{Arora:2022}, too, have suggested that multilingual LLMs, which are trained on text in many languages, may have the potential to pick up cultural values due to the diversity in languages in their training data. However, the lack of diversity within available multilingual training data may still cause multilingual LLMs to perform inconsistently across different languages and cultural contexts.  \citet{Benkler:2023} highlight that most current AI systems reflect the dominant values of the culture that produces the majority of training data and models. The authors argue that, due to this largely Western and, more specifically, English nature of the training data, LLMs have a moral bias wherein the values of W.E.I.R.D. societies are wrongfully assumed to be universal. \\
\indent Studies on AI ethics emphasize the need for models that respect cultural differences and promote equitable treatment \citep{Floridi:2018}. Accurately reflecting diverse cultural perspectives is crucial for AI systems to be fair and inclusive \citep{zowghi2023, cachat2023, karpouzis2024, Mehrabi:2021}. However, research has shown that biases embedded in training data or model design can lead to disparities in how AI systems interpret and respond to inputs from different cultural backgrounds and contexts. This variability raises questions about the universal applicability and fairness of AI systems \citep{karpouzis2024}. Studies such as those by \citet{Arora:2022} and \citet{Benkler:2023} have highlighted that LLMs may struggle to accurately represent diverse moral frameworks across different cultures. Work by \citet{Ramezani:2023}, on the other hand, shows more promising results regarding LLMs' capability of capturing cultural diversity. This divergence in findings underscores the need for further research to bridge the gap in understanding how LLMs perceive and represent moral values across diverse cultural contexts. While most studies suggest that LLMs can mirror some cultural biases present in their training data, their accuracy in representing diverse moral judgments is not yet well understood \citep{Caliskan:2017}.

\subsection{The risk of bias in LLMs} 
The training process of LLMs involves vast datasets sourced from the internet, which inherently contain societal and cultural biases. These biases can be reflected and even amplified in model outputs, leading to concerns about fairness and representation. This works as follows: LLMs learn to understand language through word embeddings, which are dense vector representations of words that capture the semantic and syntactic relationships between them. Based on the co-occurrence patterns present in large text corpora, these vectors encode contextual information about words. Through these word embeddings, LLM's can pick up biases similar to those of humans from the word associations in their training data \citep{Nemani:2024}. In short, LLM's form associations between words and concepts through word embeddings based on their co-occurrence in the training data, which may lead to biased predictions and outputs. Biased outputs from LLMs should be combated, as they can perpetuate stereotypes, reinforce prejudices, and lead to unfair treatment of certain groups. \\
\indent Various studies have documented biases in LLMs, including gender, racial, and cultural biases \citep{Bender:2021, Buolamwini:2018}. \citet{bolukbasi2016}, for example, showed that word embeddings can encode significant gender stereotypes regarding profession: the authors found notable associations stereotypically linking 'woman' with 'homemaker' and 'man' with 'computer programmer'. Moreover, \citet{Johnson:22} highlight GPT-3's stereotyping bias shown by association of the word "Muslims" with violent actions much more often than "Christians". Despite efforts to mitigate bias in LLMs \citep{mishra2024}, significant challenges remain in eliminating bias. Striving to eliminate bias is an important task as the incorporation of biased language in AI-systems can influence public opinion and decision-making processes, thereby potentially causing harm \citep{Noble:2018}.  For example, if an LLM is trained on biased data, it might generate job recommendations that favor men over women for technical roles, thus promoting gender inequality \citep{bolukbasi2016}. Similarly, GPT-3 linking Muslims to violent actions can lead to heightened social prejudice and discrimination against individuals who identify as Muslim. This biased association in AI-generated content may reinforce negative stereotypes and lead to increased stigmatization of Muslim communities. These significant societal implications of biased AI outputs underscore the importance of developing models that accurately reflect cultural diversity and moral values \citep{Zou:2018}.

\section{Datasets}

\subsection{World Values Survey}
World Values Survey \smallfootnote{https://www.worldvaluessurvey.org/WVSDocumentationWV7.jsp} collects data on people's values across cultures in a detailed way. The Ethical Values and Norms section in World Values Survey Wave 7 is the first dataset used in this study. This wave ran from 2017 to 2020 and is publicly available \citep{Haerpfer:2022}. In this segment, participants from 55 countries were surveyed on their views regarding 19 morally-related statements, such as divorce, euthanasia, political violence, and cheating on taxes. The questionnaire was translated into the primary languages spoken in each country and offered multiple response options.

\begin{figure}[h]
\centering
\includegraphics[width=0.4\textwidth]{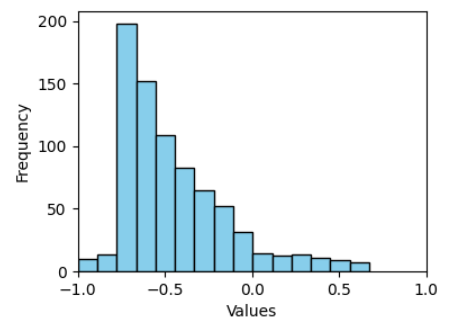}
\caption{Distribution of normalized answer values for WVS wave 7}
\label{fig1}
\end{figure}

\begin{figure}[h]
\centering
\includegraphics[width=0.35\textwidth]{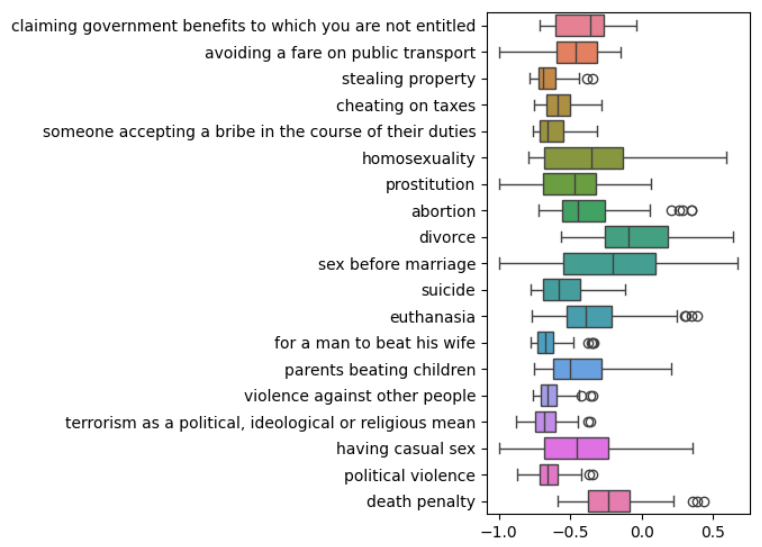}
\caption{Spread of responses across the moral topics and countries for WVS wave 7}
\label{fig2}
\end{figure}

From the original dataset, only the country name and the answer to each question were retained, which were then normalized between -1 and 1. These scores indicate how justifiable a moral value is, ranging from -1 ('never justifiable') to 1 ('always justifiable'). Normalization is applied to standardize data, enhance comparability, facilitate statistical analysis, and aid in interpretation. To calculate the moral rating for each country-moral value pair, the survey responses were averaged. This approach offers a snapshot of collective attitudes toward moral values within each country. However, it's important to note potential limitations. Averaging may oversimplify diverse viewpoints and obscure outlier perspectives. Additionally, the averaging process may obscure outliers or minority perspectives that could offer insights into the complexities of moral reasoning within a society. However, in the context of this study, averaging emerged as the most viable approach. The distribution of the aggregated and normalized answer values, as well as the spread of responses across the different moral topics, are depicted in Figures \ref{fig1} and \ref{fig2}, respectively.

\subsection{PEW 2013 Global Attitudes Survey}
The second dataset is a survey conducted by the Pew Global Attitudes Project \smallfootnote{https://www.pewresearch.org/dataset/spring-2013-survey-data/}, offering a comprehensive collection of data reflecting people's assessments of their views on current global affairs and significant contemporary issues. Undertaken in 2013, this survey offers insights into 8 morally related topics like getting a divorce or drinking alcohol, with 100 participants from each of the 39 countries contributing to the dataset. The survey questions were given in English and presented three options: 'morally acceptable,' 'not a moral issue,' and 'morally unacceptable.'

\begin{figure}[ht]
\centering
\includegraphics[width=0.4\textwidth]{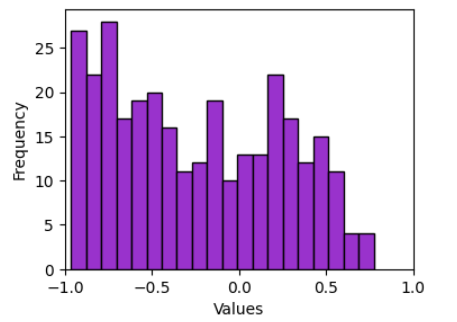}
\caption{Distribution of normalized answer values for PEW 2013}
\label{fig3}
\end{figure}

\begin{figure}[ht]
\centering
\includegraphics[width=0.4\textwidth]{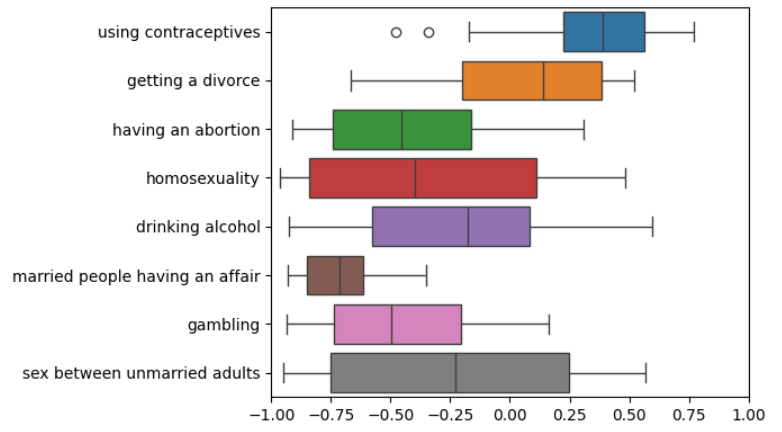}
\caption{Spread of responses across the moral topics and countries for PEW 2013}
\label{fig4}
\end{figure}

From the original dataset, only the country names and responses to questions Q84A to Q84H were retained. These responses were then normalized to range between -1 and 1. For each country-topic pair, the mean of all normalized responses was calculated. Figures \ref{fig3} and \ref{fig4} illustrate the distribution of these aggregated, normalized values and the variation in responses across different moral topics, respectively.

\section{Methodology}

\subsection{Pre-processing}
In the preprocessing of version 5 of the World Values Survey (WVS) data, the dataset was initially filtered to retain only the columns corresponding to the moral questions Q177 to Q195 and the country code (B\_COUNTRY). These questions cover a range of moral issues, such as tax cheating, accepting bribes, and attitudes towards homosexuality. Following the initial filtering, country names were assigned to each row based on the B\_COUNTRY codes using a predefined country mapping dataset. Responses with values of -1, -2, -4, and -5, which represent 'Don't know,' 'No answer,' 'Not asked in survey,' and 'Missing; Not available,' respectively, were replaced with zero. This adjustment was made to ensure that calculations, such as averaging, were not affected by non-responses. The decision to replace with 0 ensures that the structure of the dataset remains intact. It avoids introducing NaN values or leaving cells empty, which could complicate subsequent data analysis tasks such as averaging or statistical modeling. Moreover, a replacement value of 0 ensures that non-responses do not influence the computed averages or other aggregated measures artificially.  
After replacing non-response values with 0, the dataset was aggregated by country, calculating the mean response for each moral question per country. This provided a country-specific average score for each ethical issue. To enable comparisons across different countries and questions, these average scores were normalized on a scale from -1 to 1, where 1 signifies that the behavior is justifiable in every case and -1 denotes it is never justifiable. This normalization involved adjusting the mean responses, which initially ranged from 1 to 10, to fit the new scale. This step was needed for cross-national comparisons. Finally, normalized values were rounded to four decimal places to enhance clarity.

\subsection{Models}
In this study, various natural language processing (NLP) models with the same architecture type are used to explore how moral values differ across cultures, based on responses to a series of statements. To ensure a standardized basis for comparison, all models selected are transformer-based and well-suited for the task of text generation. All of the models tested have a decoder-only architecture. Models of this architecture were chosen due to their ability to generate text based on contextual prompts. This makes them suitable for the task of exploring and comparing moral values across different cultural contexts flexibly. The models used in this research originate from Hugging Face \smallfootnote{https://huggingface.co/}, a well-known provider of cutting-edge NLP models. Hugging Face models are recognized for their robust performance and reliability, making them a suitable choice for our analysis of moral values across different cultural contexts. Importantly, none of the models were trained or fine-tuned for this study, as the goal is to understand the inherent perspectives these models hold regarding moral topics without the influence of training on similar datasets.

\subsubsection{Monolingual Models}

The first part of the study involves employing two monolingual models. The first one is the \textbf{GPT-2} language model, which is primarily trained on English text. GPT-2 was chosen for its strong performance in generating coherent and contextually relevant text, as demonstrated in various studies. This model has been fine-tuned to accurately predict the probability of a word based on its context within a sentence. Its architecture and training process enable it to generate human-like text, making it a suitable choice for tasks involving nuanced language understanding \cite{Radford2019LanguageMA}.

In particular, two versions of GPT-2 were utilized to assess the influence of model size on moral understanding. The models utilized are 'GPT-2 Medium' with 355 million parameters, and 'GPT-2 Large' with 774 million parameters. All models were sourced from Hugging Face. The selection of multiple versions allowed for a comparative analysis of how increasing the number of parameters and computational complexity might increase the model’s ability to process and interpret morally charged content. Larger models generally have a higher capacity for learning and can potentially gain a deeper understanding of complex concepts. This approach provides insights into whether increased computational resources allow for a more accurate reflection of different cultural moral judgments. 

The OPT model, part of the Open Pre-trained Transformer \textbf{ (OPT)} series developed by Meta AI, is the second model included in this study. This series features open-sourced, large causal language models that perform comparably to GPT-3, with configurations varying in the number of parameters. Two such variants, the OPT-125M and the OPT-350M, are used in this analysis. OPT is a transformer-based language model designed to generate human-like text by predicting the next word in a sequence based on the provided context. Primarily trained on English text, OPT has been exposed to diverse datasets, enabling it to effectively handle a wide range of text generation tasks. This model was selected for its balance between computational efficiency and performance, providing a benchmark for comparing smaller, resource-efficient models against larger, more complex models \citep{zhang2022opt}.

\subsubsection{Multilingual Models}

The second part of the study involves employing multiple multilingual models. Using multilingual models allows for an analysis of how these models, trained on a diverse and extensive dataset, influence moral judgments across different countries compared to monolingual models. 

The first multilingual model used is the BigScience Large Open-science Open-access Multilingual Language Model, commonly known as \textbf{BLOOM}. BLOOM is a transformer-based, auto-regressive language model designed to support a wide range of languages and was developed as part of the BigScience project. It has been trained transparently on diverse datasets encompassing 46 natural and 13 programming languages, making it highly versatile and capable of generating text across various languages and contexts \citep{Scao2022BLOOMA1}. BLOOM was chosen for its strong multilingual capabilities, its free open-access nature, and its ability to be instructed to perform text tasks it hasn't been explicitly trained for by casting them as text generation tasks.

A variant of BLOOM-560M, known as BLOOMZ-560M, which also has 560 million parameters and is provided by BigScience (bigscience/bloomz-560m), was chosen since it is fine-tuned for enhanced performance on zero-shot learning tasks, making it better at generalizing to new tasks without extensive training. Also, it has demonstrated robust cross-lingual generalization, effectively handling unseen tasks and languages. Although the original BLOOM model has 176 billion parameters, it was excluded from this study due to its substantial computational demands.

The \textbf{Qwen} model, developed by the Alibaba Cloud team, was also included in this study. Qwen is another multilingual transformer-based language model trained on data in 29 languages, including English and Chinese. Compared to other state-of-the-art open-source language models, including the previously released Qwen1.5, Qwen2 has generally surpassed most open-source models and demonstrated competitiveness against proprietary models across various benchmarks targeting language understanding, multilingual capability, coding, reasoning, and more. It is available in four parameter sizes: 0.5B, 1.5B, 7B, and 72B. The 0.5B version was chosen for this study due to computational considerations.

\subsection{Calculating moral judgment scores}
In assessing whether the chosen LLMs capture patterns of divergence and agreement on topics across cultures, three methods are used: (1) comparing variances of model-generated and survey-based moral scores across countries, (2) assessing alignment between country clusters derived from model-generated moral scores and those derived from survey-based moral scores, and (3) probing LLMs with direct comparative prompts to evaluate their recognition of cultural differences and similarities in moral judgments. For all three methods, moral scores obtained by probing the LLMs with certain prompts are used. For the first two methods, moral scores are computed for all country-topic pairs by assessing the log probabilities that follow from probing the model. In probing the model, the following two prompt structures are used for all country-topic pairs: 

\begin{itemize}
    \item In \{country\} \{topic\} is \{moral\_judgment\}.
    \item People in \{country\} believe \{topic\} is \{moral\_judgment\}.
\end{itemize}

The model is prompted with the above prompts, making use of five different token pairs. Each token pair consists of two contrasting statements for the moral judgment:

\begin{itemize}
    \setlength\itemsep{-0.5em}
    \item pair1 = (always justifiable, never justifiable)
    \item pair2 = (right, wrong)
    \item pair3 = (morally good, morally bad)
    \item pair4 = (ethically right, ethically wrong)
    \item pair5 = (ethical, unethical)
\end{itemize}

For instance, when probing the model to obtain the moral score for the United States on the topic of abortion, the contrasting prompts for the first prompt structure and the first token pair are: "In the United States abortion is always justifiable" and "In the United States abortion is never justifiable". The model responds to these prompts in the form of log probabilities that indicate the model's predicted moral judgment for each prompt. The responses are then used to compute moral scores for each country-topic pair as follows: the moral score for each moral-immoral token pair is computed by subtracting the log probability for the non-moral statement from the log probability for the moral statement. The result of all token pairs is then averaged to obtain the final model-generated moral score. This is done for both prompt-styles, 'people' and 'in', and the results are averaged to obtain a final moral score for each country-topic pair, akin to the ground-truth WVS scores for each country-topic pair.  In summary, these prompts and token pairs are systematically applied across all country-topic pairs to obtain model-generated moral scores. By using contrasting token pairs and averaging results from different prompt styles, this method allows for a nuanced analysis of LLMs' reflection of moral attitudes across various topics and cultural contexts.

For the third method, the models will be probed with prompts consisting of direct comparative statements. The prompts used for this method will be described in paragraph 4.3.3. 

\subsection{Methods}

\subsubsection{Comparison of variances}
Firstly, a comparison of variances between countries’ moral scores on the given ethical topics was conducted. For each ethical topic, the cross-cultural variance in scores was calculated, identifying topics with high variance (more controversial) and low variance (more agreed upon) across cultures. This process was applied to both the WVS and PEW survey moral scores as well as the model-generated moral scores. For every survey-model pair, the two sets of variance scores (one originating from the survey moral scores, and one originating from the model-generated moral scores) were then compared and assessed for association. A strong positive association indicates that the language model effectively captures cross-country variations in moral scores, while a weak positive association or a negative association indicates the opposite. Pearson’s correlation coefficient was used to test the strength and significance of the association. This method was also used by \citet{Ramezani:2023} in assessing the ability of LLM’s in capturing which topics people across cultures tend to diverge or agree on in their moral judgment. Furthermore, the variance scores were also used to assess which topics were identified as most controversial and most agreed upon according to the surveys and the models, thereby aiming to detect any notable alignments or discrepancies.

\subsubsection{Cluster alignment}
Secondly, a clustering approach was utilized to examine the models’ abilities to capture the differences and similarities in moral judgments on ethical topics between different countries. The aim of this method was to assess the alignment between country clusterings resulting from the survey scores with country clusterings resulting from the model scores. Hereby, the model’s ability to replicate empirically observed cultural patterns of divergence and agreement could be assessed. For this method, countries were first clustered  based on their WVS moral scores for the specified ethical topics using K-means clustering. The value for hyperparameter K (i.e. the number of clusters) was chosen by running the clustering algorithm multiple times, each time with a different value for K from a prespecified range of [1,10]. The value for K that yielded the clustering result with the highest silhouette score was chosen. After having obtained the clustering result based on the WVS moral scores, K-means clustering was also used on the model-generated moral scores for countries. Since the goal is to compare how well the model clusterings align with the ground truth survey clusterings, the value for K that was chosen for clustering countries based on the survey moral scores was also used in this clustering step. The resulting two clusterings of countries (one clustering based on the ground truth moral scores, and the other clustering based on the model-generated moral scores) were then compared for alignment. Subsequently, the degree of alignment between the clusterings based on the survey scores and the model-generated scores was assessed using the Adjusted Rand Index (ARI) and Adjusted Mutual Information (AMI) metrics, which are well-known measures of the similarity between two data clusterings \citep{Nazaretsky2020, lazarenko2021}. Additionally, the average of the ARI and AMI was computed to produce a Combined Alignment Score (CAS), providing a holistic measure of alignment between the survey-based and model-generated clusterings. \\
\indent For both the WVS and PEW survey scores, the survey-based clusterings were compared with the model-based clusterings thrice: first including the moral scores on all topics, then including only the moral scores on the most controversial topics, and lastly including only the moral scores on the most agreed upon topics. By first including all moral topics, the general alignment across a broad spectrum of moral topics is assessed. By then focusing on the most controversial topics, the model's ability to capture areas where moral judgments are highly variable and contentious is examined. Lastly, by including only the topics with the highest agreement, the model's ability to recognize common moral ground across different cultures is evaluated. Using the clustering method on the three subsets allows for a comprehensive analysis of the alignment between survey-based and model-generated clusterings, thereby providing a detailed and nuanced understanding of how well the models align with human moral judgments. 

\subsubsection{Probing with direct comparative prompts}
Thirdly, the models were probed with prompts designed to directly assess their understanding of cultural differences and similarities regarding moral judgments on isolated ethical topics. Specifically, probing with direct comparative prompts was used to assess whether the models recognize that countries belonging to the same cluster (intracluster countries) are similar regarding their moral judgment on the topic in question, while countries belonging to different clusters (intercluster countries) are different in this regard. For each topic, hierarchical clustering was used to cluster countries based on their survey moral score for that topic. For this, the elbow method as described by \citet{SAPUTRA2020} was used to find the optimal number of K clusters. Once the clusters were found, the two most differing clusters for the topic were selected, i.e. the two clusters with the largest difference in their average moral score. Only the two most differing clusters were selected for further analysis. The two most differing clusters were isolated to determine whether the model recognizes the differences in moral attitudes between the most empirically divergent country clusters regarding the moral scores they ascribe to different topics. As mentioned above, this was tested by probing the model with prompts that directly include this comparison. Over the course of 50 trials, two representative countries were randomly chosen from both clusters for the generation of comparative statements. The prompt structure used was: 

\begin{itemize}
    \item Regarding the morality of \{topic\}, the judgments of people in \{country x\} and \{country y\} are \{comparative token\}
\end{itemize}

For the intracluster prompts, 'country x' and 'country y' belong to the same cluster, while they belong to different clusters for the intercluster prompts. The model was prompted with the above intra- and intercluster country comparisons, making use of three different token pairs. Each token pair consists of two contrasting terms for the comparative token:

\begin{itemize}
    \setlength\itemsep{-0.5em}
    \item pair1 = (similar, dissimilar)
    \item pair2 = (alike, unalike)
    \item pair3 = (aligned, misaligned)
\end{itemize}

After prompting the model using the above token pairs, the average log probabilities in favour of the tokens signifying that the countries are similar or different were calculated. If the average log probability in favour of 'similar' was higher than the average log probability in favour of 'different', the country-pair was labeled 'similar' in their model-estimated attitude on the topic in question, and vice versa if the average log probability in favour of 'different' was higher. The country-pairs on the topic in question were also empirically labeled 'similar' in the case of intracluster country pairs, and 'different' in the case of intercluster country pairs. The results were then evaluated using confusion matrix metrics, as well as a Chi-squared test for association, assessing how well the model-estimated labels aligned with the empirically determined cluster labels for similarity or dissimilarity.

\raggedbottom 

\section{Results}

\subsection{Comparison of variances}  

\begin{table}[H]
    \centering
    \begin{tabular}{lll}  
        \toprule
        \emph{Model} & \emph{r} & \emph{p}\\
        \midrule
        GPT-2 Medium          & -0.195   & 0.424      \\
        GPT-2 Large   & -0.115   & 0.640     \\
        OPT-125       & -0.035        & 0.887             \\
        QWEN          & -0.200        & 0.413             \\
        BLOOM   &-0.118 &0.631  
        \\
        \bottomrule
    \end{tabular}
    \caption{Correlation results for WVS dataset topic variances and model-generated scores topic variances} 
    \label{tab:table1} 
\end{table}

The analysis reveals that the moral score variances of the WVS survey and the model are weakly negatively correlated. The weak negative correlations observed between the WVS variance and model variation are not statistically significant, implying that the model's variance does not correspond significantly to the WVS variance. There is no evidence from this analysis to suggest that the variance in WVS scores is related to the variance in the moral scores generated by the LLMs for these topics. This suggests that the examined models do not accurately capture the variability in moral judgments across different cultures. In short, the weak negative correlations, combined with their insignificance, highlight limitations of the models in their capability to capture the intercultural nuances of moral dimensions as reflected in the WVS survey.

\begin{table}[H]
    \centering
    \small
    \begin{tabular}{lll}
        \toprule
        \emph{Source} & \emph{Mean moral score} & \emph{Mean variance} \\
        \midrule
        WVS & -0.575995 & 0.075163 \\
        BLOOM & 0.473576 & 0.003785 \\
        OPT-125 & 0.104104 & 0.011770 \\
        QWEN & 0.241515 & 0.021157 \\
        GPT-2 Large & 0.322543 & 0.014663 \\
        GPT-2 Medium & 0.411380 & 0.012844 \\
        \bottomrule
    \end{tabular}
    \caption{Mean moral and variance scores by source}
    \label{tab:combined_moral_variation_scores}
\end{table}

Generally, the models ascribe more positive moral scores and lower variance scores to the various WVS topics than is empirically observed in the survey, as can be seen in table \ref{tab:combined_moral_variation_scores}, showcasing the models' tendencies to mistakenly judge most topics as more morally accepted and uniformly agreed upon globally than they are in reality. When looking at the WVS variance scores, we can identify the topics that are empirically most controversial (highest variance scores) and most agreed upon (lowest variance scores). These topics are shown in tables \ref{tab:controversialWVS} and \ref{tab:agreedWVS}, respectively. The most controversial and agreed upon topics according to the models can be found in the appendix, in tables \ref{tab:controversial_topicsWVSgpt2-m} to \ref{tab:agreed_topicsWVSbloom}. It can be observed that the models do not correctly grasp which topics are most controversial and agreed upon across cultures. Most notably, we observe that sex before marriage and homosexuality are by far the two most controversial topics according to the WVS survey. The models, however, generally do not capture this extremity. Saliently, Qwen, and BLOOM even include one of these two topics in the top 3 most agreed upon topics, as can be observed in tables \ref{tab:agreed_topicsWVSqwen} and \ref{tab:agreed_topicsWVSbloom}. 

\begin{table}[H]
    \centering
    \small
    \begin{tabular}{lll}
        \toprule
        \emph{Topic} & \emph{Variance} \\
        \midrule
        Sex before marriage  & 0.219 \\
        Homosexuality  &0.209 \\
        Euthanasia  &0.1264 \\
    \end{tabular}
    \caption{Top 3 most controversial WVS topics}
    \label{tab:controversialWVS}
\end{table}

\begin{table}[H]
    \centering
    \small
    \begin{tabular}{lll}
        \toprule
        \emph{Topic} & \emph{Variance} \\
        \midrule
          Stealing property&0.015  \\
          Violence against other people&0.015 \\
          For a man to beat his wife&0.018 \\
    \end{tabular}
    \caption{Top 3 most agreed upon WVS topics}
    \label{tab:agreedWVS}
\end{table}

\begin{figure*}
    \centering
    \includegraphics[width=\linewidth]{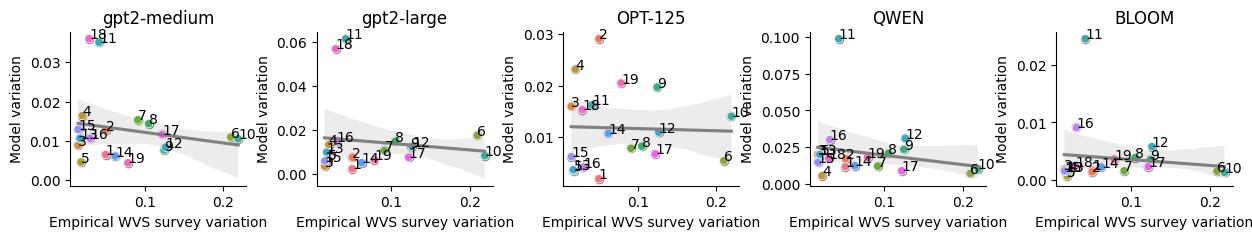}
    \caption{Comparison between the degrees of cultural diversities and shared tendencies in the empirical moral ratings and language-model inferred moral scores for WVS}
    \label{fig:enter-label}
\end{figure*}


\begin{figure*}
    \centering
    \includegraphics[width=\linewidth]{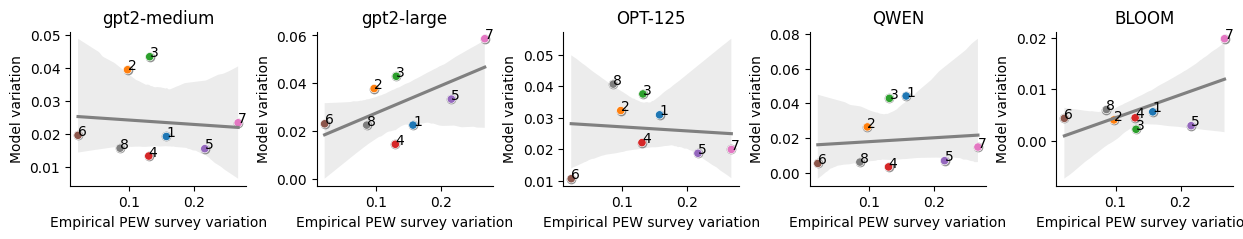}
    \caption{Comparison between the degrees of cultural diversities and shared tendencies in the empirical moral ratings and language-model inferred moral scores for PEW}
    \label{fig:enter-label}
\end{figure*}

Compared to the correlation scores for the WVS dataset, the correlation scores between the PEW dataset moral score variance and the model-generated moral score variance are slightly more favorable, as can be inferred from table \ref{tab:PEWcomparisonvariance}. Notably, a moderate to strong positive correlation can be observed for GPT-2 Medium and BLOOM. This suggests that these models are better at capturing the cultural variance on the topics present in the PEW survey. While the correlation scores for the PEW dataset include moderate to strong positive correlations for GPT-2 Large and BLOOM, none of the correlations reach statistical significance. 

\begin{table}[H]
    \centering
    \begin{tabular}{lll}  
        \toprule
        \emph{Model} & \emph{r} & \emph{p} \\
        \midrule
        GPT-2 Medium          & -0.090   & 0.832    \\
        GPT-2 Large   &0.617  &0.103    \\
        OPT-125       &-0.095    &0.822     \\
        QWEN          &0.102    &0.811     \\
        BLOOM   &0.608 &0.110
        \\
        \bottomrule
    \end{tabular}
    \caption{Correlation results for PEW dataset topic variances and model-generated scores topic variances} 
    \label{tab:PEWcomparisonvariance} 
\end{table}

\begin{table}[H]
    \centering
    \small
    \begin{tabular}{lll}
        \toprule
        \emph{Source} & \emph{Mean moral score} & \emph{Mean variance} \\
        \midrule
        PEW & -0.244369 & 0.138406 \\
        BLOOM & 0.246118 & 0.006169 \\
        OPT-125 & 0.247627 & 0.026601 \\
        QWEN & 0.221191 & 0.018898 \\
        GPT-2 Large & 0.160102 & 0.031917 \\
        GPT-2 Medium & 0.226518 & 0.023776 \\
        \bottomrule
    \end{tabular}
    \caption{Mean moral and variance scores by source}
    \label{tab:combined_scores}
\end{table}

Similar to the WVS variance scores, the models generally ascribe higher moral scores and lower variance scores to the PEW topics than those empirically observed, as can be seen in table \ref{tab:combined_scores}. Again, this underlines the models' inability to discern intercultural variability, inaccurately comprehending most topics as more generally morally acceptable across cultures. The same two topics that were identified as most controversial cross-culturally according to the WVS data, i.e. sex before marriage and homosexuality, also come forth as most controversial in the PEW data. As can be inferred from tables \ref{tab:agreed_topicsPEWgpt2_medium}, \ref{tab:agreed_topicsPEWopt125} and \ref{tab:agreed_topicsPEWbloom} one or both of these topics are again incorrectly seen as one of the most agreed on cross-culturally by some models, namely by GPT-2 Medium, OPT-125 and BLOOM.


\begin{table}[H] 
    \centering
    \small
    \begin{tabular}{lll}
        \toprule
        \emph{Topic} & \emph{Variance} \\
        \midrule
          Sex between unmarried adults &0.268 \\
          Homosexuality &0.216 \\
          Drinking alcohol &0.157 \\
    \end{tabular}
    \caption{Top 3 most controversial PEW topics}
    \label{tab:my_label}
\end{table}

\begin{table}[H]
    \centering
    \small
    \begin{tabular}{lll}
        \toprule
        \emph{Topic} & \emph{Variance} \\
        \midrule
          Married people having an affair&0.021  \\
          Using contraceptives &0.086 \\
          Gambling&0.097 \\

    \end{tabular}
    \caption{Top 3 most agreed upon PEW topics}
    \label{tab:my_label}
\end{table}

\begin{table*} 
    \centering
    \small
    \begin{tabularx}{\linewidth}{l l l *{4}{>{\raggedright\arraybackslash}X}}
        \toprule
        \textbf{Model} & \textbf{WVS variance} & \textbf{WVS mean} & \textbf{Model variance} & \textbf{Model mean} & \textbf{Topic} & \textbf{Variance difference} \\
        
        GPT-2 Medium & 0.219 & -0.244 & 0.011 & 0.465 & sex before marriage & 0.208 \\
        GPT-2 Medium & 0.209 & -0.396 & 0.011 & 0.577 & homosexuality & 0.198 \\
        GPT-2 Medium & 0.126 & -0.430 & 0.008 & 0.481 & euthanasia & 0.118 \\
        GPT-2 Medium & 0.125 & -0.150 & 0.008 & 0.217 & divorce & 0.117 \\
        GPT-2 Medium & 0.122 & -0.452 & 0.012 & 0.371 & having casual sex & 0.110 \\
        
        GPT-2 Large & 0.219 & -0.244 & 0.008 & 0.454 & sex before marriage & 0.211 \\
        GPT-2 Large & 0.209 & -0.396 & 0.018 & -0.086 & homosexuality & 0.192 \\
        GPT-2 Large & 0.122 & -0.452 & 0.008 & 0.470 & having casual sex & 0.114 \\
        GPT-2 Large & 0.126 & -0.430 & 0.013 & 0.261 & euthanasia & 0.114 \\
        GPT-2 Large & 0.125 & -0.150 & 0.012 & 0.121 & divorce & 0.112 \\
        
        OPT-125 & 0.219 & -0.244 & 0.014 & 0.475 & sex before marriage & 0.205 \\
        OPT-125 & 0.209 & -0.396 & 0.005 & 0.255 & homosexuality & 0.204 \\
        OPT-125 & 0.126 & -0.430 & 0.011 & 0.013 & euthanasia & 0.115 \\
        OPT-125 & 0.122 & -0.452 & 0.007 & 0.093 & having casual sex & 0.115 \\
        OPT-125 & 0.125 & -0.150 & 0.020 & -0.261 & divorce & 0.105 \\
        
        QWEN & 0.219 & -0.244 & 0.010 & 0.415 & sex before marriage & 0.209 \\
        QWEN & 0.209 & -0.396 & 0.007 & 0.466 & homosexuality & 0.202 \\
        QWEN & 0.122 & -0.452 & 0.009 & 0.177 & having casual sex & 0.113 \\
        QWEN & 0.125 & -0.150 & 0.024 & -0.042 & divorce & 0.101 \\
        QWEN & 0.126 & -0.430 & 0.031 & -0.115 & euthanasia & 0.095 \\
        
        BLOOM & 0.219 & -0.244 & 0.001 & 0.662 & sex before marriage & 0.218 \\
        BLOOM & 0.209 & -0.396 & 0.002 & 0.865 & homosexuality & 0.208 \\
        BLOOM & 0.124 & -0.150 & 0.004 & 0.569 & divorce & 0.121 \\
        BLOOM & 0.126 & -0.429 & 0.006 & 0.712 & euthanasia & 0.121 \\
        BLOOM & 0.122 & -0.452 & 0.002 & 0.422 & having casual sex & 0.120 \\
        \bottomrule
    \end{tabularx}
    \caption{Topics with largest difference in survey and model variance - WVS}
    \label{tab:wvs_models_data}
\end{table*}

\begin{table*} 
    \centering
    \small
    \begin{tabularx}{\linewidth}{l l l *{4}{>{\raggedright\arraybackslash}X}}
        \toprule
        \textbf{Model} & \textbf{PEW variance} & \textbf{PEW mean} & \textbf{Model variance} & \textbf{Model mean} & \textbf{Topic} & \textbf{Variance difference} \\
        
        GPT-2 Medium & 0.268 & -0.219 & 0.023 & 0.044 & sex between unmarried adults & 0.244 \\
        GPT-2 Medium & 0.216 & -0.342 & 0.016 & 0.641 & homosexuality & 0.201 \\
        GPT-2 Medium & 0.157 & -0.234 & 0.019 & 0.142 & drinking alcohol & 0.138 \\
        
        GPT-2 Large & 0.268 & -0.219 & 0.059 & -0.138 & sex between unmarried adults & 0.209 \\
        GPT-2 Large & 0.216 & -0.342 & 0.033 & -0.188 & homosexuality & 0.183 \\
        GPT-2 Large & 0.157 & -0.234 & 0.023 & 0.210 & drinking alcohol & 0.135 \\
        
        OPT-125 & 0.268 & -0.219 & 0.020 & 0.512 & sex between unmarried adults & 0.248 \\
        OPT-125 & 0.216 & -0.342 & 0.019 & 0.570 & homosexuality & 0.198 \\
        OPT-125 & 0.157 & -0.234 & 0.031 & 0.187 & drinking alcohol & 0.126 \\
        
        QWEN & 0.268 & -0.219 & 0.015 & 0.494 & sex between unmarried adults & 0.253 \\
        QWEN & 0.216 & -0.342 & 0.007 & 0.562 & homosexuality & 0.209 \\
        QWEN & 0.130 & -0.405 & 0.004 & 0.130 & having an abortion & 0.127 \\
        
        BLOOM & 0.268 & -0.219 & 0.020 & 0.374 & sex between unmarried adults & 0.248 \\
        BLOOM & 0.216 & -0.342 & 0.003 & 0.843 & homosexuality & 0.213 \\
        BLOOM & 0.157 & -0.234 & 0.006 & 0.159 & drinking alcohol & 0.152 \\
        \bottomrule
    \end{tabularx}
    \caption{Topics with largest difference in survey and model variance - PEW}
    \label{tab:pew_models_data}
\end{table*}

\subsection{Cluster alignment}  

\subsubsection{All topics}

\begin{table}[H] 
        \begin{tabular}{ l l l l}  
        \toprule
        \emph{Model} & \emph{ARI}  & \emph{AMI} & \emph{CAS} \\
            \midrule
                  GPT-2 Medium        &-0.012                 &-0.002
                              &-0.007
                              \\
                  GPT-2 Large  &0.028                 &0.04  
                               &0.034 
                               \\
                  OPT-125      &-0.073                 &0.037
                               &-0.018
                               \\
                  QWEN         &0.291                 &0.138 
                               &0.215
                    
                  \\
                  BLOOM        &0.015
                               &-0.011
                               &0.002
                    \\
                  \bottomrule
             \hline
        \end{tabular}
    \caption{Result of alignment between clusters based on WVS moral scores and model-generated moral scores for all topics} 
    \label{tab:table1} 
\end{table}

The cluster alignment scores for most models are very low. The model that performs best is QWEN, with alignment scores much higher than those of the other models, suggesting that the moral scores generated by QWEN yield a clustering that, out of the tested models, best reflects the empirical moral scores for all WVS topics. 

\begin{table}[H] 
        \begin{tabular}{ l l l l}  
        \toprule
        \emph{Model} & \emph{ARI}  & \emph{AMI} & \emph{CAS} \\
            \midrule
                  GPT-2 Medium        &0.087                 &0.068
                              &0.078
                              \\
                  GPT-2 Large  &0.129                 &0.123  
                               &0.126 
                               \\
                  OPT-125      &0.129                 &0.123
                               &0.126
                               \\
                  QWEN         &-0.019                 &0.065 
                               &0.023
                  \\
                  BLOOM        &0.008
                               &-0.004
                               &0.002
                    \\
                  \bottomrule
             \hline
        \end{tabular}
    \caption{Result of alignment between clusters based on PEW moral scores and model-generated moral scores for all topics} 
    \label{tab:table1} 
\end{table}

The alignment scores for clustering based on PEW moral scores and model-generated moral scores vary significantly across the models tested. GPT-2 Large and OPT-125 show a modest degree of alignment, with identical ARI scores of 0.129 and AMI scores of 0.123, indicating some degree of similarity in clustering patterns. GPT-2 Medium alignment scores are not very high, but scores for QWEN and BLOOM are even lower, with BLOOM exhibiting the lowest alignment.

\subsubsection{Most controversial topics}

\begin{table}[H] 
        \begin{tabular}{ l l l l}  
        \toprule
        \emph{Model} & \emph{ARI}  & \emph{AMI} & \emph{CAS} \\
            \midrule
                  GPT-2 Medium        &-0.015                 &-0.011
                              &-0.013
                              \\
                  GPT-2 Large  &-0.012                 &0.023 
                               &0.005 \\
                               OPT-125      &-0.021                 &0.017
                               &-0.002
                  \\                  QWEN         &-0.014                 &-0.018 
                               &-0.016
                    \\
                
                  BLOOM        &-0.015
                               &-0.011
                               &-0.013
                    \\
                  \bottomrule
             \hline
        \end{tabular}
    \caption{Result of alignment between clusters based on WVS moral scores and model-generated moral scores for most controversial topics} 
    \label{tab:table1} 
\end{table}

The alignment scores for clustering based on WVS moral scores and model-generated moral scores reveal consistent negative values across all models tested. This indicates a lack of agreement in clustering patterns between the models and the WVS dataset for the most controversial topics. GPT-2 Medium, GPT-2 Large, QWEN, OPT-125, and BLOOM all exhibit negative ARI scores. The AMI scores are negative for all models except for GPT-2 Large and OPT-125. This suggests divergent cluster structures compared to the WVS moral scores for all models, with GPT-2 Large showing the best performance relatively while QWEN performs worst.

\begin{table}[H] 
        \begin{tabular}{ l l l l}  
        \toprule
        \emph{Model} & \emph{ARI}  & \emph{AMI} & \emph{CAS} \\
            \midrule
                  GPT-2 Medium        &-0.026                 &-0.019
                              &-0.022
                              \\
                  GPT-2 Large  &0.093                 &0.081  
                               &0.087 
                                \\
                  OPT-125      &0.131                 &0.14
                               &0.136                               \\
                  QWEN         &-0.006                 &0.073 
                               &0.033
                  \\
                  BLOOM        &0.009
                               &0.006
                               &0.007
                    \\
                  \bottomrule
             \hline
        \end{tabular}
    \caption{Result of alignment between clusters based on PEW moral scores and model-generated moral scores for most controversial topics} 
    \label{tab:table1} 
\end{table}

The alignment scores show varied results across the models tested. GPT-2 Medium demonstrates negative alignment scores across all metrics, indicating significant divergence in cluster structures compared to the PEW dataset. In contrast, GPT-2 Large, OPT-125, and BLOOM exhibit positive alignment scores, suggesting some degree of agreement in clustering patterns with the PEW moral scores. Notably, OPT-125 shows the highest alignment scores among all models tested, indicating a stronger correspondence with the moral judgments reflected in the PEW dataset.

\subsubsection{Most agreed upon topics}
\begin{table}[H] 
        \begin{tabular}{ l l l l}  
        \toprule
        \emph{Model} & \emph{ARI}  & \emph{AMI} & \emph{CAS} \\
            \midrule
                  GPT-2 Medium        &0.079                 &0.01
                              &0.044
                              \\
                  GPT-2 Large  &-0.019                 &-0.014  
                               &-0.016 
                                       \\
                  OPT-125      &0.12                 &0.038
                               &0.079                            \\
                  QWEN         &-0.005                 &-0.017 
                               &-0.011

                  \\
                  BLOOM        &-0.03
                               &-0.012
                               &-0.021
                    \\
                  \bottomrule
             \hline
        \end{tabular}
    \caption{Result of alignment between clusters based on WVS moral scores and model-generated moral scores for most agreed upon topics} 
    \label{tab:table1} 
\end{table}

The alignment scores scores show mixed results across the models tested for the most agreed upon topics. GPT-2 Medium and OPT-125 demonstrate positive alignment scores across all metrics, suggesting some agreement in cluster structures with the WVS dataset. In contrast, GPT-2 Large, QWEN and BLOOM exhibit negative alignment scores, indicating divergence in clustering patterns compared to the WVS moral scores and suggesting minimal agreement in clustering for these topics.
\begin{table}[H] 
        \begin{tabular}{ l l l l}  
        \toprule
        \emph{Model} & \emph{ARI}  & \emph{AMI} & \emph{CAS} \\
            \midrule
                  GPT-2 Medium        &0.057                 &0.045
                              &0.051
                              \\
                  GPT-2 Large  &0.028                 &0.02  
                               &0.024 
                               \\
                                
                  OPT-125      &0.035                 &0.051
                               &0.043            \\      QWEN         &-0.02                 &-0.016 
                               &-0.018
                  \\
                  BLOOM        &0.006
                               &0.004
                               &0.005
                    \\
                  \bottomrule
             \hline
        \end{tabular}
    \caption{Result of alignment between clusters based on PEW moral scores and model-generated moral scores for most agreed upon topics} 
    \label{tab:table1} 
\end{table}

The alignment scores reveal generally modest alignment across the models tested for the most agreed upon PEW topics. GPT-2 Medium and GPT-2 Large exhibit moderate alignment scores, suggesting some degree of similarity in cluster structures with the PEW dataset. OPT-125 also shows moderate alignment. In contrast, QWEN and BLOOM demonstrate very low or slightly negative alignment scores, indicating a divergence in clustering patterns compared to the PEW moral scores.

\subsection{Probing with direct comparative prompts} 

\begin{table}[H] 
        \begin{tabular}{p{1.2cm} l l l l }  
        \toprule
        \emph{Model} &
        \emph{Accuracy} & \emph{Precision}  & \emph{Recall} & \emph{F1}    \\
            \midrule
                  GPT-2 Medium        &0.485&0.488&0.336&0.398\\
                  GPT-2 Large   & 0.509& 0.508& 0.946& 0.661\\
                  OPT-125 &0.502&0.51&0.461&0.484\\
        QWEN    &0.5&0.504&0.831&0.628\\
        BLOOM   &0.495&0.543&0.026&0.05\\
                  \bottomrule
             \hline
        \end{tabular}
    \caption{Confusion matrix scores for direct probing - WVS} 
    \label{tab:confusionWVS} 
\end{table}

The confusion matrix scores for direct probing based on WVS data as displayed in table \ref{tab:confusionWVS} show varying performance across the models tested. While all models achieve similar accuracy scores of around 0.5, indicating comparable overall prediction performance, their precision, recall, and F1 scores differ significantly. 

GPT-2 Large and QWEN stand out with high recall scores of 0.946 and 0.831, respectively, caused by their ability to correctly identify instances of one of the two classes. OPT-125 and GPT-2 Medium, which achieve similar accuracy, show lower recall and F1 scores. Although the F1 scores are meagre, they are higher than the F1 score for BLOOM, which demonstrates the lowest overall performance with an accuracy score below 0.5 and notably lower precision, recall, and F1 scores. This suggests challenges in the model's predictive capabilities.

\begin{table}[H] 
        \begin{tabular}{ l l l }  
        \toprule
        \emph{Model} & \emph{$\chi^2$}  & \emph{p} \\
            \midrule
                  GPT-2 Medium          &8.38&0.004**\\
                  GPT-2 Large   & 1.491&0.222\\
                  OPT-125   &0.338&0.561\\
                  QWEN  &1.416&0.234\\
                  BLOOM     &1.279&0.258\\
                  \bottomrule
             \hline
        \end{tabular}
    \caption{Chi-squared test for correlation between scores and model judgment regarding recognition of intra- and intercluster (dis)similarity of moral values - WVS} 
    \label{tab:table1} 
\end{table}

For GPT-2 Medium, the p-value is less than 0.01, indicating a statistically significant association between WVS scores and the model's judgments. For the other models, the Chi-squared test results for association between WVS scores and model judgment regarding recognition of intra- and intercluster (dis)similarity of moral values indicate no statistically significant associations across the models tested. These results suggest that GPT-2 Medium might be better at understanding or reflecting differences and similarities in human moral judgments as captured by the WVS compared to the other models.

\begin{table}[H] 
        \begin{tabular}{p{1.2cm} l l l l }  
        \toprule
        \emph{Model} &
        \emph{Accuracy} & \emph{Precision}  & \emph{Recall} & \emph{F1}    \\
            \midrule
                  GPT-2 Medium        &0.495&0.494&0.402&0.444\\
                  GPT-2 Large   &0.495&0.497&0.954&0.654\\
                  OPT-125 &0.506&0.506&0.48&0.493\\
        QWEN    &0.493&0.495&0.694&0.578\\
        
        BLOOM   &0.497&0.326&0.006&0.011\\
                  \bottomrule
             \hline
        \end{tabular}
    \caption{Confusion matrix scores for direct probing - PEW} 
    \label{tab:table1} 
\end{table}

Again, all models achieve an accuracy of around 0.5, indicating moderate overall prediction performance. GPT-2 Large shows an accuracy of 0.495 with strong recall (0.954), indicating its ability to correctly identify a high proportion of positive instances. Similarly, QWEN demonstrates an accuracy of 0.493 and a recall score of 0.694, showing similar performance to GPT-2 Large in recall. While GPT-2 Medium's accuracy is on par with the other models, its precision, recall and F1 scores suggest room for improvement. OPT-125 achieves an accuracy of 0.506 with balanced precision (0.506) and recall (0.480), but its F1 score is slightly lower at 0.493. BLOOM performs with an accuracy of 0.497, but its precision, recall, and F1 scores are notably lower, indicating limitations in correctly predicting both positive and negative instances.

\begin{table}[H] 
        \begin{tabular}{ l l l }  
        \toprule
        \emph{Model} & \emph{$\chi^2$}  & \emph{p} \\
            \midrule
                  GPT-2 Medium          &0.418&0.518\\
                  GPT-2 Large   &3.325&0.068\\
                  OPT-125   &0.609&0.435\\
                  QWEN  &1.017&0.313\\
                  BLOOM     &4.599&0.032*\\
                  \bottomrule
             \hline
        \end{tabular}
    \caption{Chi-squared test for correlation between scores and model judgment regarding recognition of intra- and intercluster (dis)similarity of moral values - PEW} 
    \label{tab:table1} 
\end{table}

The Chi-squared test results for correlation between WVS scores and model judgment regarding recognition of intra- and intercluster (dis)similarity of moral values show varied findings across the models tested.

GPT-2 Medium, GPT-2 Large, OPT-125 and QWEN exhibit p-values greater than 0.05, indicating no statistically significant association. This suggests that these models do not recognize empirically observed intra- and intercluster (dis)similarity well on the basis of direct comparative prompts.

In contrast, BLOOM exhibits a statistically significant p-value of 0.014, indicating a meaningful correlation between its judgments on moral (dis)similarity and the WVS labels. This might suggest that BLOOM captures certain nuances or patterns related to moral (dis)similarity that are statistically detectable despite its poor classification metrics. However, since the p-value does not specify the direction of the association, a significant p-value could mean that BLOOM's judgments align with or differ from the WVS scores in a consistent manner in either a positive or negative direction.  

\section{Discussion and conclusion}
The findings of this study shed light on the capability of large language models to accurately capture cultural diversity and common tendencies across different moral topics. The investigation utilized multiple methodologies that were based on probing LLMs with prompts derived from the World Values Survey (WVS) and PEW datasets, focusing on a range of moral topics.

\subsection{Comparison of variance}
The correlation analysis between model-generated moral scores and empirical survey data revealed mixed results. For the PEW dataset, GPT-2 Large and BLOOM demonstrated moderate to strong alignment in capturing cultural variations. The fact that the largest model (GPT-2 Large) and the largest multilingual model (BLOOM) performed best may suggest that model size and multilinguality have a positive effect on models' ability to grasp patterns of cultural diversity, which would be in line with previous work from \citet{du-2024} and \citet{Arora:2022}. However, the correlations did not reach statistical significance and therefore no strong claims can be made. Moreover, model performance shows high variability, with weak negative correlations observed for both GPT-2 Large and BLOOM when comparing their variances with the WVS moral score variances. The other models performed weakly and variably in both the PEW and WVS moral score variance comparisons. Furthermore, the models struggled to accurately identify the most controversial and agreed on topics. In fact, some of the models incorrectly categorized (one of) the two most controversial topics as among the most agreed on. The variable and low overall performance could be attributed to the fact that the complexity and nuance of moral values across different cultural contexts may not be fully captured by the models' training data.

\subsection{Cluster alignment}
The clustering alignment results further emphasized the variability in model performance. Overall, GPT-2 Large and OPT-125 showed better alignment with empirical moral scores from both datasets quite consistently, suggesting their relative proficiency in clustering countries based on moral attitudes. However, other models, most notably BLOOM, exhibited lower alignment scores, indicating shortcomings in their ability to mirror the clustering patterns observed in the survey data. These results suggest that the models fall short in grasping cultural patterns regarding moral judgments, which is in line with the findings from the previous method. Thus, while GPT-2 Large and OPT-125 generally show better alignment with empirical moral scores across various topics, the variability in model performance underscores the challenges in accurately capturing the complexities of moral attitudes across different cultural contexts. Overall, the clusterings based on the model scores do not faithfully capture the cultural patterns observed in the clusterings derived from the survey scores.

\subsection{Probing with direct comparative prompts} 
Direct probing with comparative prompts provided additional insights into the models' understanding of moral differences between culturally distinct groups. In general, performance is low as the scores are no higher or even slightly lower than random chance. GPT-2 Large and QWEN stood out with higher accuracy and recall scores, indicating their better performance in distinguishing moral differences between the most divergent clusters identified by the survey data. Upon further inspection, however, it became clear that GPT-2 Large and QWEN almost always predict the same class, which does not signify a proper understanding of inter-cultural differences and similarities. If we disregard the performance of GPT-2 Large and QWEN due to the fact that they always predict the same class, GPT-2 Medium and OPT-125 exhibit the most balanced performance across the remaining models. BLOOM exhibited the lowest performance metric scores, suggesting challenges in discerning nuanced moral judgments across cultures. Notably, despite its low overall performance, BLOOM's judgments were found to be statistically associated with the judgments based on the PEW dataset through a Chi-squared test. This suggests that there may be some alignment between BLOOM's outputs and the moral judgments reflected in the PEW dataset. However, it is important to note that this statistical association does not necessarily imply a meaningful understanding or accurate representation of moral differences between cultures.   

\subsection{Conclusion}   
In conclusion, the study underscores the importance of rigorous evaluation methodologies when assessing LLMs' ability to understand and reflect cultural diversity in moral judgments. The tested models seem to propagate a homogenized view on cross-cultural moral values, identifying most topics as cross-culturally agreed on as more morally acceptable than empirically observed. Thereby, the models generally seem to reflect a rather liberal view, in line with the autonomy-endorsing values found in W.E.I.R.D. societies \citep{graham2016}. It has been established in the literature that exclusively English training data plays a big part in the embedding of homogenous W.E.I.R.D. values and, thereby, cultural bias in LLMs \citep{Benkler:2023}. This could lead one to believe that multilingual LLMs are the answer to mitigating bias in LLMs \citep{Arora:2022}. However, this study could not find convincing evidence to suggest that multilingual models are better at truthfully capturing cultural diversities in moral judgments than monolingual models. Similarly, while model size could be considered another factor influencing model performance due to its potential to enhance computational capacity and capture more complex patterns \citep{du-2024}, its impact was not found to be convincing in the carried out analyses. It can be concluded that this study found no remarkable differences between the tested models in their success, regardless of multilinguality or model size. Overall, the models examined show variable performance and generally exhibit low success in aligning with empirical moral data from global surveys. Thus, ongoing research and development are needed to enhance their accuracy and reliability in diverse cultural settings. Addressing these challenges is crucial for ensuring the ethical integrity and societal impact of AI technologies in the context of global applications. 

\section{Limitations}  

Despite the insights gained from this study, several limitations must be acknowledged.

Firstly, the ground truth moral survey datasets (WVS and PEW), while comprehensive, inevitably oversimplify the complex reality of human values and behaviors. This limitation arises from the inherent constraints of survey-based research, where nuanced and multifaceted moral and ethical beliefs are condensed into predefined categories and responses. The survey datasets primarily capture explicit moral values, potentially overlooking implicit or subconscious values that significantly influence human behavior. Moreover, the individual survey responses were averaged to obtain aggregated moral scores for countries, which further simplifies cultural moral perspectives. As a result, this analysis might not fully take into account the deeper, underlying moral frameworks that individuals operate within.

Secondly, the study tested only a few models, which restricts the ability to make generalized statements about how well LLMs reflect the empirically observed variations and similarities in moral judgments across cultures. The limited scope of model testing means that the conclusions drawn may not be broadly applicable across different analytical frameworks or contexts.

Thirdly, the probing in this analysis was conducted using a limited set of specific prompts, and the style and formulation of these prompts can significantly influence the outcomes \citep{Wang2024}. This highlights the importance of prompt engineering in shaping the responses and suggests that different prompt styles could yield varying results, thereby introducing a degree of variability and potential bias into the findings.

Lastly, in method 3 (probing with direct comparative prompts), representatives from clusters were randomly chosen for the analysis due to computational limitations. This random selection process may not accurately represent the full diversity within each cluster, potentially skewing the results and limiting the robustness of the conclusions drawn from this method.

The limitations described above underscore the need for cautious interpretation of the findings and suggest avenues for future research to address these constraints. Further studies could benefit from a more comprehensive approach, including a broader range of models, diverse prompt engineering strategies, and more exhaustive methods for direct comparative probing.


\bibliography{main}
\bibliographystyle{acl_natbib}





\section{Appendix}

\subsection{Most controversial WVS topics according to models}

\begin{table}[H]
    \centering
    \small
    \begin{tabular}{p{3.75cm}p{2cm}l}
        \toprule
        \emph{Topic} & \emph{Model} & \emph{Variance} \\
        \midrule
        Political violence & GPT-2 Medium & 0.036 \\
        Suicide & GPT-2 Medium & 0.035 \\
        Cheating on taxes & GPT-2 Medium & 0.016 \\
        \bottomrule
    \end{tabular}
    \caption{Top 3 most controversial WVS topics according to GPT-2 Medium}
    \label{tab:controversial_topicsWVSgpt2-m}
\end{table}

\begin{table}[H]
    \centering
    \small
    \begin{tabular}{p{3.75cm}p{2cm}l}
        \toprule
        \emph{Topic} & \emph{Model} & \emph{Variance} \\
        \midrule
        Suicide & GPT-2 Large & 0.062 \\
        Political violence & GPT-2 Large & 0.057 \\
        Homosexuality & GPT-2 Large & 0.018 \\
        \bottomrule
    \end{tabular}
    \caption{Top 3 most controversial WVS topics according to GPT-2 Large}
    \label{tab:controversial_topics_WVSgpt2-l}
\end{table}

\begin{table}[H]
    \centering
    \small
    \begin{tabular}{p{3.75cm}p{2cm}l}
        \toprule
        \emph{Topic} & \emph{Model} & \emph{Variance} \\
        \midrule
        Avoiding a fare on public transport & OPT-125 & 0.029 \\
        Cheating on taxes & OPT-125 & 0.023 \\
        Death penalty & OPT-125 & 0.021 \\
        \bottomrule
    \end{tabular}
    \caption{Top 3 most controversial WVS topics according to OPT-125}
    \label{tab:controversial_topicsWVSopt125}
\end{table}

\begin{table}[H]
    \centering
    \small
    \begin{tabular}{p{3.75cm}p{2cm}l}
        \toprule
        \emph{Topic} & \emph{Model} & \emph{Variance} \\
        \midrule
        Suicide & QWEN & 0.099 \\
        Terrorism as a political, ideological or religious tactic & QWEN & 0.030 \\
        Euthanasia & QWEN & 0.031 \\
        \bottomrule
    \end{tabular}
    \caption{Top 3 most controversial WVS topics according to QWEN}
    \label{tab:controversial_topicsWVSqwen}
\end{table}

\begin{table}[H]
    \centering
    \small
    \begin{tabular}{p{3.75cm}p{2cm}l}
        \toprule
        \emph{Topic} & \emph{Model} & \emph{Variance} \\
        \midrule
        Suicide & BLOOM & 0.025 \\
        Terrorism as a political, ideological or religious tactic & BLOOM & 0.009 \\
        Euthanasia & BLOOM & 0.006 \\
        \bottomrule
    \end{tabular}
    \caption{Top 3 most controversial WVS topics according to BLOOM}
    \label{tab:controversial_topicsWVSbloom}
\end{table}

\subsection{Most agreed on WVS topics according to models}

\begin{table}[H]
    \centering
    \small
    \begin{tabular}{p{3.75cm}p{2cm}l}
        \toprule
        \emph{Topic} & \emph{Model} & \emph{Variance} \\
        \midrule
        Death penalty & GPT-2 Medium & 0.004 \\
        Accepting a bribe in the course of duty & GPT-2 Medium & 0.005 \\
        Parents beating children & GPT-2 Medium & 0.006 \\
        \bottomrule
    \end{tabular}
    \caption{Top 3 most agreed on WVS topics according to GPT-2 Medium}
    \label{tab:agreed_topicsWVSgpt2-m}
\end{table}

\begin{table}[H]
    \centering
    \small
    \begin{tabular}{p{3.75cm}p{2cm}l}
        \toprule
        \emph{Topic} & \emph{Model} & \emph{Variance} \\
        \midrule
        Claiming government benefits to which you are entitled & GPT-2 Large & 0.002 \\
        Stealing property & GPT-2 Large & 0.004 \\
        Parents beating children & GPT-2 Large & 0.005 \\
        \bottomrule
    \end{tabular}
    \caption{Top 3 most agreed on WVS topics according to GPT-2 Large}
    \label{tab:agreed_topicsWVSgpt2-l}
\end{table}

\begin{table}[H]
    \centering
    \small
    \begin{tabular}{p{3.75cm}p{2cm}l}
        \toprule
        \emph{Topic} & \emph{Model} & \emph{Variance} \\
        \midrule
        Claiming government benefits to which you are entitled & OPT-125 & 0.002 \\
        Someone accepting a bribe in the course of duty & OPT-125 & 0.003 \\
        For a man to beat his wife & OPT-125 & 0.004 \\
        \bottomrule
    \end{tabular}
    \caption{Top 3 most agreed on WVS topics according to OPT-125}
    \label{tab:agreed_topicsWVSopt125}
\end{table}

\begin{table}[H]
    \centering
    \small
    \begin{tabular}{p{3.75cm}p{2cm}l}
        \toprule
        \emph{Topic} & \emph{Model} & \emph{Variance} \\
        \midrule
        Cheating on taxes & QWEN & 0.006 \\
        Homosexuality & QWEN & 0.007 \\
        Having casual sex & QWEN & 0.009 \\
        \bottomrule
    \end{tabular}
    \caption{Top 3 most agreed on WVS topics according to QWEN}
    \label{tab:agreed_topicsWVSqwen}
\end{table}

\begin{table}[H]
    \centering
    \small
    \begin{tabular}{p{3.75cm}p{2cm}l}
        \toprule
        \emph{Topic} & \emph{Model} & \emph{Variance} \\
        \midrule
        Someone accepting a bribe in the course of duty & BLOOM & 0.001 \\
        Sex before marriage & BLOOM & 0.001 \\
        Avoiding a fare on public transport & BLOOM & 0.001 \\
        \bottomrule
    \end{tabular}
    \caption{Top 3 most agreed on WVS topics according to BLOOM}
    \label{tab:agreed_topicsWVSbloom}
\end{table}

\subsection{Most controversial PEW topics according to models}

\begin{table}[H]
    \centering
    \small
    \begin{tabular}{p{3.75cm}p{2cm}l}
        \toprule
        \emph{Topic} & \emph{Model} & \emph{Variance} \\
        \midrule
        Getting a divorce & GPT-2 Medium & 0.043 \\
        Gambling & GPT-2 Medium & 0.039 \\
        Sex between unmarried adults & GPT-2 Medium & 0.023 \\
        \bottomrule
    \end{tabular}
    \caption{Top 3 most controversial PEW topics according to GPT-2 Medium}
    \label{tab:controversial_topicsPEWgpt2_medium}
\end{table}

\begin{table}[H]
    \centering
    \small
    \begin{tabular}{p{3.75cm}p{2cm}l}
        \toprule
        \emph{Topic} & \emph{Model} & \emph{Variance} \\
        \midrule
        Sex between unmarried adults & GPT-2 Large & 0.059 \\
        Getting a divorce & GPT-2 Large & 0.043 \\
        Gambling & GPT-2 Large & 0.038 \\
        \bottomrule
    \end{tabular}
    \caption{Top 3 most controversial PEW topics according to GPT-2 Large}
    \label{tab:controversial_topicsPEWgpt2_large}
\end{table}

\begin{table}[H]
    \centering
    \small
    \begin{tabular}{p{3.75cm}p{2cm}l}
        \toprule
        \emph{Topic} & \emph{Model} & \emph{Variance} \\
        \midrule
        Using contraceptives & OPT-125 & 0.041 \\
        Getting a divorce & OPT-125 & 0.038 \\
        Gambling & OPT-125 & 0.032 \\
        \bottomrule
    \end{tabular}
    \caption{Top 3 most controversial PEW topics according to OPT-125}
    \label{tab:controversial_topicsPEWopt125}
\end{table}

\begin{table}[H]
    \centering
    \small
    \begin{tabular}{p{3.75cm}p{2cm}l}
        \toprule
        \emph{Topic} & \emph{Model} & \emph{Variance} \\
        \midrule
        Drinking alcohol & QWEN & 0.044 \\
        Getting a divorce & QWEN & 0.043 \\
        Gambling & QWEN & 0.027 \\
        \bottomrule
    \end{tabular}
    \caption{Top 3 most controversial PEW topics according to QWEN}
    \label{tab:controversial_topicsPEWqwen}
\end{table}

\begin{table}[H]
    \centering
    \small
    \begin{tabular}{p{3.75cm}p{2cm}l}
        \toprule
        \emph{Topic} & \emph{Model} & \emph{Variance} \\
        \midrule
        Sex between unmarried adults & BLOOM & 0.020 \\
        Using contraceptives & BLOOM & 0.006 \\
        Drinking alcohol & BLOOM & 0.006 \\
        \bottomrule
    \end{tabular}
    \caption{Top 3 most controversial PEW topics according to BLOOM}
    \label{tab:controversial_topicsPEWbloom}
\end{table}

\subsection{Most agreed on PEW topics according to models}

\begin{table}[H]
    \centering
    \small
    \begin{tabular}{p{3.75cm}p{2cm}l}
        \toprule
        \emph{Topic} & \emph{Model} & \emph{Variance} \\
        \midrule
        Having an abortion & GPT-2 Medium & 0.013 \\
        Homosexuality & GPT-2 Medium & 0.016 \\
        Using contraceptives & GPT-2 Medium & 0.016 \\
        \hline
    \end{tabular}
    \caption{Top 3 most agreed on PEW topics according to GPT-2 Medium}
    \label{tab:agreed_topicsPEWgpt2_medium}
\end{table}

\begin{table}[H]
    \centering
    \small
    \begin{tabular}{p{3.75cm}p{2cm}l}
        \toprule
        \emph{Topic} & \emph{Model} & \emph{Variance} \\
        \midrule
        Having an abortion & GPT-2 Large & 0.015 \\
        Using contraceptives & GPT-2 Large & 0.023 \\
        Drinking alcohol & GPT-2 Large & 0.023 \\
        \bottomrule
    \end{tabular}
    \caption{Top 5 most agreed on PEW topics according to GPT-2 Large}
    \label{tab:agreed_topicsPEWgpt2_large}
\end{table}

\begin{table}[H]
    \centering
    \small
    \begin{tabular}{p{3.75cm}p{2cm}l}
        \toprule
        \emph{Topic} & \emph{Model} & \emph{Variance} \\
        \midrule
        Married people having an affair & OPT-125 & 0.011 \\
        Homosexuality & OPT-125 & 0.019 \\
        Sex between unmarried adults & OPT-125 & 0.020 \\
        \bottomrule
    \end{tabular}
    \caption{Top 3 most agreed on PEW topics according to OPT-125}
    \label{tab:agreed_topicsPEWopt125}
\end{table}

\begin{table}[H]
    \centering
    \small
    \begin{tabular}{p{3.75cm}p{2cm}l}
        \toprule
        \emph{Topic} & \emph{Model} & \emph{Variance} \\
        \midrule
        Having an abortion & QWEN & 0.004 \\
        Married people having an affair & QWEN & 0.006 \\
        Using contraceptives & QWEN & 0.006 \\
        \bottomrule
    \end{tabular}
    \caption{Top 3 most agreed on PEW topics according to QWEN}
    \label{tab:agreed_topicsPEWqwen}
\end{table}

\begin{table}[H]
    \centering
    \small
    \begin{tabular}{p{3.75cm}p{2cm}l}
        \toprule
        \emph{Topic} & \emph{Model} & \emph{Variance} \\
        \midrule
        Getting a divorce & BLOOM & 0.002 \\
        Homosexuality & BLOOM & 0.003 \\
        Gambling & BLOOM & 0.004 \\
        \bottomrule
    \end{tabular}
    \caption{Top 3 most agreed on PEW topics according to BLOOM}
    \label{tab:agreed_topicsPEWbloom}
\end{table}

\section*{Acknowledgements}
I would like to thank my supervisors, Dr. Ayoub Bagheri \& Hadi Mohammadi, for their guidance and support throughout the research process. I hereby also express my gratitude to Evi and Lourenço for being great collaborators in this project. Moreover, I want to thank Dewi and David, for the much needed stargazing that was done in between drafts in Zeeland, and Giulia, for exerting her organizational capacities and booking us our weekly study rooms at Drift. I also want to thank Siebe, for his perseverance and positivity which have inspired me beyond measure this year. Lastly, I want to thank all my wonderful friends, family and my dear Petr for the love, support and confidence they continuously instill in me and the joy they bring to my life. 

\end{document}